# EFFECTIVE FACE LANDMARK LOCALIZATION VIA SINGLE DEEP NETWORK

*Zongping Deng[1,2], Ke Li[1], Qijun Zhao[1,2], Yi Zhang[2] and Hu Chen[1,2,3]*

[1]National Key Laboratory of Fundamental Science on Synthetic Vision
[2]School of Computer Science, Sichuan University, Chengdu, China, 610065
[3]huchen@scu.edu.cn

## ABSTRACT

In this paper, we propose a novel face alignment method using single deep network (SDN) on existing limited training data. Rather than using a max-pooling layer followed one convolutional layer in typical convolutional neural networks (CNN), SDN adopts a stack of 3 layer groups instead. Each group layer contains two convolutional layers and a max-pooling layer, which can extract the features hierarchically. Moreover, an effective data augmentation strategy and corresponding training skills are also proposed to overcome the lack of training images on COFW and 300-W datasets. The experiment results show that our method outperforms state-of-the-art methods in both detection accuracy and speed.

*Index Terms*—Face Alignment, Single Deep Network, SDN, Layer Groups

## 1. INTRODUCTION

Face landmark localization aims to locate facial landmarks such as eyes and noses automatically, it plays a very important role in the tasks for face alignment, pose estimation and face recognition. Though great success has been achieved in this field, robust facial landmark detection still remains a formidable problem in in the presence of severe occlusion and large pose variations. Besides, real-world and complex facial landmark detection is formidable challenge in practical applications.

Face landmark localization methods are usually restrained by pose, lighting, expression and severe occlusion. The popular approaches include template fitting approaches, regression-based methods, and deep learning based skills [1, 2, 3, 4, 5, 7, 8, 9, 10, 11, 21]. Besides, there are some other else methods for face alignment using 3D methods [22, 23]. Generally, the face landmark localization is treated and solved as a single and independent problem[14, 15]. However, some methods think that face landmark detection is not a standalone problem and believe auxiliary attributes [2, 3, 21] can improve the detection accuracy.

This paper still treats the face landmark detection as a single and independent problem. Different from traditional methods, deep learning convolutional neural networks (DCNN) is used to local the face landmarks. DCNN-based method usually needs a very high amount of data to get a good model. Unfortunately, the need of huge data for face alignment is hard to reach. To solve this problem, an effective data augmentation strategy and corresponding training skills are applied on the proposed single deep network and existing limited data. The SDN with proposed data augmentation strategy and train skills outperforms the state-of-the-art methods on COFW and 300-W datasets, and has high speed for detection on an Intel Core i5 CPU with average 13ms.

Inspired by method [16, 17], SDN is designed using a stack of two convolutional layers straightway without spatial max-pooling in between. This prove very more effective than two groups of one convolutional network layer followed by a max-pooling layer.

**The main contributions of our work are:**

1 - A novel network for face alignment, named single deep network (SDN). SDN is more robust to severe occlusion with simple method and easier skills. Moreover, SDN is a fast method and thus suitable for real applications.

2 - An effective data augmentation strategy and corresponding training skills are designed to overcome the lack of training images. The detail is described in Sect. 3.

## 2. RELATED WORK

Early works on face landmark detection includes Active Contours Models [6], Template Matching, Active Shape Models (ASM) [14], Appearance Models (AAM) [15]. Recently, deep learning based methods [1, 2, 3, 5, 11, 13] have shown their advantages for face landmark localization. Sun et al. [1] applies a DCNN based method to handle the problem of face landmark detection via three-level cascaded convolutional networks (totally 23 CNNs). Zhang et al. [5] uses successive auto-encoder networks for face alignment (**CFAN**) by multiple convolutional networks. Hou et al. [12] presents a novel cascade multi-channel convolutional neural networks (**CMC-CNN**), several networks are jointly used for face alignment. Zhang et al. [2, 3] uses the same network structure with [1] for face landmark localization with multi-tasks which named **TCDCN**. Shao et al. [11] adopts a deep network learning deep representation from coarse to fine for face alignment (**CFT**). Although these methods above achieve a great improment for face alignment, they are complicated an hard to realize, and demand huge data. To simple the CNN based method, Deng et al. [13] proposes a new single deep network (**SDN**), which uses only one simple deep CNN and achieves the state-of-the-art results for 5 points detection. However, same to other CNN based methods, it also demands high amount of data to achieve better performance.

## 3. PROPOSED METHOD

### 3.1 The proposed single deep network

The network contains a stack of 3 layer groups to extract the features hierarchically, followed by two fully-connected layers and the output indicates the position of the 2*N landmarks.

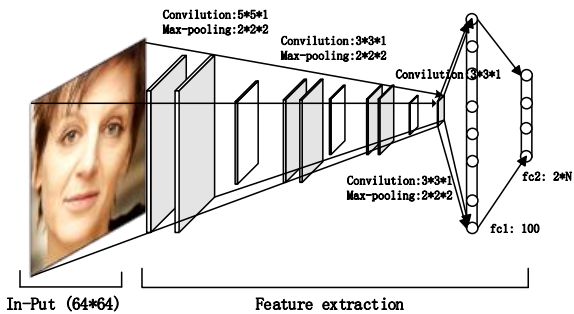

Fig.1 the structure of proposed network, the gray regions are convolutional layers and the white ones are Max-polling layers, the last two layers are fully-connected unit.

Different from typical CNN based methods [1, 2, 3, 5, 13], our method adopts a stack of two convolutional layers $n \times n$ (without spatial max-pooling in between) which has an effective receptive field of $(2n − 1) \times (2n − 1)$ and two non-linear rectification layers, and that makes the exacted feature more discriminative than that of a single convolutional layer.

Since our experiments are conducted on the normalized datasets (minus its own mean value as the final input data), the hyperbolic tangent function (tanh) is used as activation function, which can be defined as:

$$tanh(z) = \frac{e^z - e^{-z}}{e^z + e^{-z}}$$

The loss is formulated as follows:

$$loss = \frac{1}{2N} \sum_{i=1}^{N} \left\| S_p^i - S_g^i \right\|_2$$

where N is the batch-size. $S_g^i$ is the ground true shape, $S_p^i$ is the regression target.

### 3.2 Data argument and Training strategy
**Data argument.** Generally speaking, translation, rotation, mirroring, and stretch are used for augment training data. These methods can create many times data from given database. However, the limited original data leads to robust to different scenario hard. To solve this problem, an effective method to make much more effective data from limited data is proposed. We design three different stages to generate different data from the original face bounding boxes.

**Stage 1:** Face bounding boxes are augmented with different sizes and the expansion ratio from 0.1 to 0.5 according to the inter–pupil distance, by extending width and height larger than original boxes. After that, these augmented face boxes are rotated with the angles from -50 to +50. The boxes without the landmarks in the box region are discarded. Finally, all the images generated with two steps above are mirrored. From the three augment step above, about 60 times data are obtained from limited given data. For example, the Helen training dataset can be extended from 2k to 120k. The augmented data is used for the prepared stage training. However, the detection isn't accurate enough (Fig.3 shows the result). Because the training data have large variation range from original face bounding box, while the input testing face bounding box, which comes from the face detection tool, is constant.

**Stage 2:** The process is similar to stage 1. Face bounding boxes are augmented with different sizes with smaller extended ratio, which is from 0.1 to 0.3, and slighter angles, whose arrange is from -20 to 20. Note that the angles are difference with stage 1. The step size of angle here is 5 while the step size of angle stage 1 is 3. Moreover, the original images are stretched to get more information. Finally, 80k data are extended from Helen training data which are different from stage 1.

**Stage 3**: This is the final stage for augmenting training data. First, the best trained model from stage 2 is used to select the hard examples from training data. Here errors larger than 0.02 in training data are considered as hard sample. The hard examples are then augmented via sight rotated angles from -10 to 10 with step size 2. All the extended images are also mirrored. Finally, about 20k images are obtained from stage 3 training.

**Training strategy:** After argumenting data above, enough data are obtained for training SDN with the deep learning framework Caffe [20]. Firstly, training data from stage 1 is used to train proposed network. The learning policy is set to be "fixed", and learning rate is set as 0.001. Secondly, the model is fine-tuned using the training data from stage 2. Different from stage 1, the learning policy is set as "step" with the step-size is 20,000 and gamma is 0.1. That means the learning rate will be reduced every 20,000 iterations. Finally, the model from stage 2 is fine-tuned with the training data from stage 3. The "inv" learning policy is used with the gamma is set as 0.00001 and the power is equal to 0.75. The learning rate changes can define as follows:

$$lr = base\_lr \qquad (1)$$

$$lr = base\_lr \times gamma\textasciicircum(floor(iter/step)) \quad (2)$$
$$lr = base\_lr \times (1 + gamma \times iter)\textasciicircum(-power) \quad (3)$$

The formula (1), (2), (3) is the presentation of **fixed**, **step**, **inv** policy respectively, where $base\_lr$ means the given learning rate. The training loss output of training are shown in the Fig.2. Fig.3 shows the performance of each stage with the accumulative error curve on Helen dataset [19].

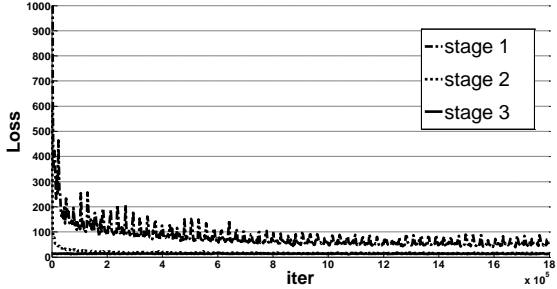

Fig.2 The loss output in each stage.

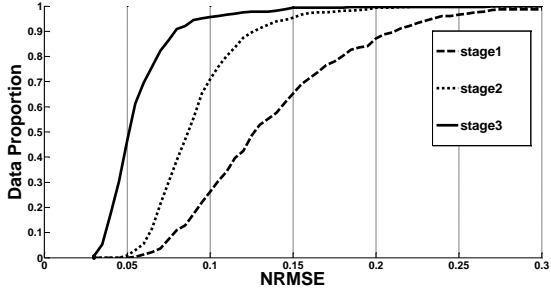

Fig.3 The accumulative error curve of each stage on Helen dataset.

### 3.3 Discussion

**Comparison with SDN [13].** Both SDN[13] and proposed method use the same network structure. The work of SDN[13] is for 5 points only with huge training data. Although its results have achieved stage-of-the-art in four public dataset, its good performance may be benefit from the big training data. The performance with the same training data as that of compared methods is not evaluated and discussed. In this paper, the same training data is used for evaluated, our model is trained using proposed data augmentation strategy and corresponding improved training skills with the training data from specific training set.

**Comparison with CFT [11].** [11] and proposed method use similar network structure, they carry their experiments with a good initialization and search optimal model smoothly. Compared to our proposal, their method depends on the principal subset and the number of selected landmarks strongly. That makes the training and detection more complicated. In the contrast, picking out principal landmarks subset to initialize network are not needed in proposed method and the experiments show that our method reaches better performance.

## 4. EXPERIMENTS

our experiments are conducted on COFW, Helen and 300-w with the same training data and testing data of [3] and [4]. For each dataset, the inter-ocular distance normalized error averages (**NRMSE**) are repoted over all landmarks, which is similar with [4] and other previous works. Moreover, failure rate are also reported. Any error above 10% is considered to be a failure. Finnally, the detection speed (**FPS**) in the same condition on each dataset is evaluated. NRMSE and failure rate are defined as follows:

$$NRMSE = \frac{\frac{1}{n}\sum_{i=1}^{n}\left\|S_{p,i} - S_{g,i}\right\|_2}{\left\|l - r\right\|_2}$$

$$failure\ rate = \frac{f_N}{N} \times 100\%$$

where $n$ is landmarks number, $l$ and $r$ are the position of the left eye corner and right eye corner, respectively. $f_N$ means the failure number of testing images, while $N$ is the total number of testing images.

**Evaluation on COFW [4].** [4] contains 1007 images annotated with 29 landmarks which are collected from the web. The training set consists of 845 LFPW face images and 500 COFW images, the faces in these images have expressions and large pose. The protocol is the same as that in [3] and [4] on COFW dataset. Some traditional method such as [7], [8]., are also compared with the proposal here. Tab. 1 shows our method performs much better than other methods except [11]. However, our method is the fastest method in all above. Note that our method is much faster than [11] and [3], which are also CNN based method.

Tab.1 The results on COFW

| method | NRMSE | FPS |
|---|---|---|
| ESR[7] | 11.2 | 4 |
| SDM[8] | 11.4 | - |
| RCPR[4] | 8.5 | 3 |
| CFT[11] | **6.33** | 15 |
| TCDCN[3] | 8.05 | 56 |
| OURS | 7.76 | **77** |

'-' denotes no report.

**Evaluation on Helen [19].** [19] contains 2000 images for training and 330 images for testing. We used the images which annotated 68 landmarks provided by [18]. The protocol is the same as that in [3] and [4] on on Helen dataset. The faces of this data set are high resolution and nearly frontal. Although our method uses down-sampled 64*64 gray images as input, it still has better performance. Tab.2 shows the results compared with [7], [8], [4], [5], [11], [10], and [3]. Note that the accuracy of our method is a little lower than results of the [3] which reported on its paper. However, our method is much better on mean error and failure rate when we use its demo that given by the website[1] on Helen, and the result is shown in Fig.4.

Tab.2 The results on Helen

| method | NRMSE | FPS | failure rate |
|---|---|---|---|
| ESR[7] | 7.10 | 2 | 13 |
| SDM[8] | 5.50 | - | - |
| RCPR[4] | 5.93 | 6 | 8 |
| CFAN[5] | 5.53 | - | - |
| CFSS[10] | 4.74 | - | - |
| CFT[11] | 4.86 | 15 | - |
| TCDCN[3] | **4.60** | 56 | - |
| OURS | 4.83 | **77** | **4.26** |

'-' denotes no report.

Tab.3 The results on 300-W

| method | common | challenging | Full |
|---|---|---|---|
| ESR[7] | 5.28 | 17.00 | 7.58 |
| SDM[8] | 5.57 | 15.40 | 7.50 |
| RCPR[4] | 6.18 | 17.26 | 8.35 |
| CFAN[5] | 5.50 | 16.78 | 7.69 |
| CFSS[10] | **4.73** | 9.98 | 5.76 |
| CMC-CNN[12] | 4.91 | 12.03 | 6.30 |
| CFT[11] | 4.82 | 10.06 | 6.62 |
| TCDCN[3] | 4.80 | 8.60 | 5.54 |
| OURS | 4.77 | **8.32** | **5.47** |

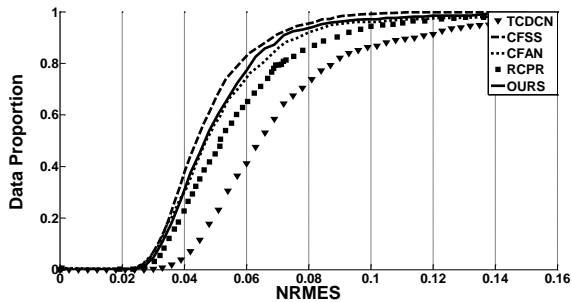

Fig. 4 The accumulative error curve for 68-pts on common set of 300-W[2].

**Evaluation on 300-W [18].** [18] is collected from existing databases such as AFW, Helen, LFPW, and IBUG. The training set contains the all images of AFW, the training set of Helen, the training set of LFPW with 3148 images totally. The testing set contains the testing set of Helen, the testing set of LFPW, and all 135 images of IBUG. The testing dataset is divided into two parts, the common subset (the testing set of Helen, and the testing data set of LFPW, totally 554 images) and the challenging subset (the IBUG [18] data set including 135 images). In the same protocol as [3, 4], the proposed method is compared with [7], [8], [4], [5], [11], [10], [12] and [3]. The results are showed in Tab.3. Moreover, the performance of accumulative error curve on the common set is also shown in Fig.4. Our method performs much better than most compared methods and a little lower than [10] on commom subset. Note that our proposal performs much better than [10] on challenging subset and total testing set. Furthermore, the method achieves average 13ms per-frame on a single core i5 CPU and 2ms on GTX760 GPU, which is much faster than that of [10](40ms). Fig. 5 shows the example results on the above dataset.

## 5. CONCLUSION

This paper presents a novel method for face landmark localization using single deep network. Profitting from the data argument strategy and corresponding training skills, our proposed method is simple and much faster than other compared methods, and acchieve the state-of-the-art performance on COFW and 300-W. The proposed method takes raw pixels as input and is very efficient. Compared with TCDCN and other methods, our method is easy to realize and outperformance. Consquently, it's suitable for real application.

## 6. ACKNOWLEDGMENT

This work is supported by the National Natural Science Foundation of China (No.61202160, NO.61302028, 61671312) and the National Key Scientific Instrument and Equipment Development Projects of China (No. 2013YQ49087904).

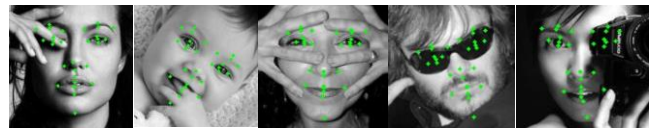

(a) Example results on COFW

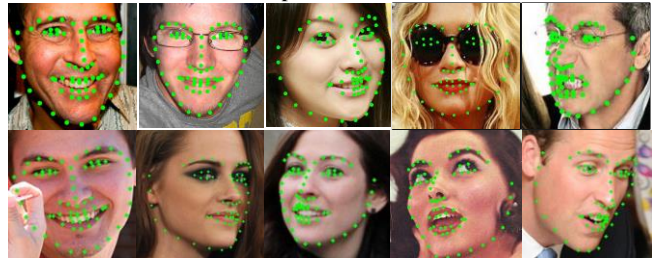

(b) Example results on 300-w
Fig. 5 Example results in our Experiments

---

[1] http://mmlab.ie.cuhk.edu.hk/projects/TCDCN.html.
[2] The data of TCDCN gets from its open source.